\def\year{2019}\relax
\documentclass[letterpaper]{article} 
\usepackage{aaai19}  
\usepackage{times}  
\usepackage{helvet}  
\usepackage{courier}  
\usepackage{amsfonts}
\usepackage{amssymb}
\usepackage{amsmath,bm}
\usepackage{graphicx}
\usepackage{algorithm}
\usepackage{algorithmic}
\usepackage{multirow}
\usepackage{amsmath}
\usepackage{xcolor}
\usepackage{subfigure}
\usepackage{breqn}
\usepackage{url}
\usepackage{epstopdf}

\begin{document}
%
\title{Ethically Aligned Opportunistic Scheduling for Productive Laziness}
\author{
Han Yu$^{1,2,3}$, Chunyan Miao$^{1,2,3}$, Yongqing Zheng$^{4,5}$, Lizhen Cui$^{4}$, Simon Fauvel$^{2}$ \textnormal{and} Cyril Leung$^{6}$\\
$^1$School of Computer Science and Engineering, Nanyang Technological University (NTU), Singapore\\
$^2$Joint NTU-UBC Research Centre of Excellence in Active Living for the Elderly (LILY), NTU, Singapore\\
$^3$Alibaba-NTU Singapore Joint Research Institute, Singapore\\
$^4$School of Software Engineering, Shandong University, Jinan, Shandong, China\\
$^5$Shanda Dareway Software Co. Ltd, Jinan, Shandong, China\\
$^6$Department of Electrical and Computer Engineering, The University of British Columbia, Vancouver, BC, Canada\\
Corresponding Authors: han.yu@ntu.edu.sg, clz@sdu.edu.cn
}
\maketitle
\begin{abstract}
In artificial intelligence (AI) mediated workforce management systems (e.g., crowdsourcing), long-term success depends on workers accomplishing tasks productively and resting well. This dual objective can be summarized by the concept of productive laziness. Existing scheduling approaches mostly focus on efficiency but overlook worker wellbeing through proper rest. In order to enable workforce management systems to follow the IEEE Ethically Aligned Design guidelines to prioritize worker wellbeing, we propose a distributed \textit{Computational Productive Laziness} (CPL) approach in this paper. It intelligently recommends personalized work-rest schedules based on local data concerning a worker's capabilities and situational factors to incorporate opportunistic resting and achieve superlinear collective productivity without the need for explicit coordination messages. Extensive experiments based on a real-world dataset of over 5,000 workers demonstrate that CPL enables workers to spend 70\% of the effort to complete 90\% of the tasks on average, providing more ethically aligned scheduling than existing approaches.
\end{abstract}

\section{Introduction}
In today's world, artificial intelligence (AI) has been employed to manage large-scale workforces such as crowdsourcing systems \cite{Miao-et-al:2016,Michelucci-Dickinson:2016,Pan-et-al:2016AAAI}. For example, in DiDi Chuxing, AI technologies are deployed to dynamically match drivers to tasks in order to enhance operation efficiency \cite{Xu-et-al:2018KDD}. Human workers become fatigued or bored over long sessions of work, which can cause inefficacy \cite{Leiter-Maslach:2015}. A recent study found that short breaks significantly improve workers' motivation while maintaining the quality of work \cite{Dai-et-al:2015}. However, existing AI approaches for workforce management in crowdsourcing mostly do not explicitly incorporate rest into their recommendations \cite{Chai-et-al:2017}.

From an ethically aligned design perspective \cite{IEEE:2018}, it is desirable to incorporate breaks into scheduling approaches in order to prioritize workers' wellbeing. Nevertheless, as highlighted in \cite{Yu-et-al:2018IJCAI}, ethically aligned AI approaches need to be designed such that they balance the concern for human wellbeing while still achieve their design objectives. In this paper, we set out to address this challenge by proposing an ethically aligned opportunistic scheduling approach that can achieve ``productive laziness'' - \textit{Computational Productive Laziness} (CPL).

Originally conceptualized in social sciences literature, productive laziness \cite{Whillans-et-al:2017} is a rule-of-thumb guideline on how workers should approach their work in order to achieve work-life balance. The general idea is to work when you are highly efficient, and rest otherwise. CPL coordinates workers to work when situational factors induce high efficiency or demand working, and rest when they do not. Thus, it is important to identify factors influencing workers' productivity and the urgency of work. We formulate the problem of achieving productive laziness in large-scale systems as a multi-objective constrained optimization problem. Following the Lyapunov optimization-based framework \cite{Yu-et-al:2016SciRep}, we analyse the interaction dynamics involved in a workforce management system and derive a novel \textit{Work-Rest Index (WRI)} which expresses the interplay among four aspects affecting worker effort output:
\begin{enumerate}
    \item \textit{Situational factors}: the current workload pending in each worker's backlog, and how long they have been pending (as tracked by the system);
    \item \textit{Worker performance}: each worker's productivity based on past observation data;
    \item \textit{System-level preference}: the emphasis, given by the system operators, on achieving high collective productivity compared to allowing for workers to be more rested; and
    \item \textit{Personal preference}: each worker's mood at a given time which is used to infer the level of productivity that can be expected from the worker in the immediate future following the happy-productive worker thesis \cite{Oswald-et-al:2015}.
\end{enumerate}

Based on this index, CPL dynamically determines the timing and amount of work a worker should perform in order to conserve the collective effort output while maintaining a high level of collective productivity in the system. It can be implemented in a distributed manner as a personal scheduling agent with a computational time complexity of $O(1)$. This enables it to effectively support the need for real-time scheduling for large-scale workforce management. CPL allows human values to be codified and algorithmically guide the recommendations by the scheduling agent to balance worker wellbeing and system throughput. Through extensive numerical experiments based on a large-scale real-world dataset containing over 5,000 workers' performance characteristics, CPL is shown to significantly outperform alternative approaches, consistently achieving superlinear collective productivity \cite{Sornette-et-al:2014}.

\section{Related Work}
Existing AI-powered workforce management approaches \cite{Lee-et-al:2015} can be divided into two main categories: (1) the direct approaches and (2) the indirect approaches.

Some direct approaches seek to balance the division of labour among workers in a situation-aware manner through data-driven deliberation \cite{Mason-Watts:2011,Yu-et-al:2013IJCAI,Dai-et-al:2013AIJ,Tran-Thanh-et-al:2014AIJ,Yu-et-al:2015AAAI,Yu-et-al:2016SciRep,Grossmann-et-al:2017,Yu-et-al:2017SciRep}. Others design reputation and/or incentive mechanisms to motivate workers to work harder \cite{Yu-et-al:2011AAMAS,Mao-et-al:2013,Yu-et-al:2013a,Tran-Thanh-et-al:2014,Miao-et-al:2016,Zeng-et-al:2017}. They generally leave it up to the workers to plan their rest breaks. There are approaches which implicitly limit how long a worker can continuously work by setting a budget on the number of tasks they are allocated \cite{Chen-et-al:2015IJCAI,Zenonos-et-al:2016}. However, these approaches do not opportunistically take advantage of periods when a worker a highly productive (e.g., periods of good mood in our approach). Such ad hoc planning may not help the system maintain high collective productivity. With CPL, a desirable balance between worker wellbeing and collective productivity can be achieved.

Indirect approaches \cite{Morris-et-al:2012} seek to induce good mood among workers in order to improve their productivity. This is based on the happy-productive worker thesis \cite{Oswald-et-al:2015}, which suggests that good (bad) mood improves (hampers) productivity. They also do not explicitly recommend resting to workers.

A recent work \cite{Yu-et-al:2017ICA} is starting to explore how to opportunistically schedule rest. However, it does not account for how long tasks have been pending in workers' backlog and can lead to schedules which do not make business sense. Without considering the urgency of the tasks, it is vulnerable to workers misreporting their mood. With the conceptual queue technique developed in this paper, CPL considers workers' wellbeing, system objectives and situational factors in a more holistic manner, and is better able to deal with misbehaving workers trying to game the system.

\section{The Proposed \textnormal{CPL} Scheduling Approach}
The system model in this paper consists of a set of $N$ workers and a set of $M$ tasks at any given time slot $t$.
\begin{itemize}
    \item \textit{Workers} are associated with personal profiles. Each profile contains information on a worker $i$'s task backlog queue $q_{i}(t)$ and his maximum productivity $\mu_{i}^{\max}$ which indicates the maximum workload he can complete in a given time slot $t$. The granularity of $t$ can be adjusted to fit any given system's requirements. The ground truth of $\mu_{i}^{\max}$ may not be directly observable. With analytics tools such as Turkalytics \cite{Heymann-Hector:2011}, it can be tracked and estimated statistically.
    \item \textit{Tasks} are associated with task profiles. The most important information for our purpose is the task deadline which is expressed in terms of number of time slots since task creation before which a given task must be completed.
\end{itemize}
The number of workers and tasks available in a given system at different time slots may differ.

We model the dynamics of a worker $i$'s task backlog queue following previous research \cite{Yu-et-al:2013IJCAI,Yu-et-al:2015AAAI}:
\begin{equation}
q_{i}(t+1)=\max[0,q_{i}(t)+\lambda_{i}(t)-\mu_{i}(t)] \label{eq:Q}
\end{equation}
where $\lambda_{i}(t)$ is the number of new tasks delegated to worker $i$ during time slot $t$; and $\mu_{i}(t)$ is the number of tasks completed by worker $i$ during time slot $t$. Based on the `happy-productive worker' thesis which suggests that productivity is positively correlated to mood \cite{Oswald-et-al:2015}, $\mu_{i}(t)$ can be expressed as a function of $i$'s current mood and how much effort he spends on the tasks:
\begin{equation}
\mu_{i}(t)=\mu(\xi_{i}(t),m_{i}(t))=\lfloor\xi_{i}(t)m_{i}(t)\mu_{i}^{\max}\rfloor \label{eq:mu1}
\end{equation}
where $\xi_{i}(t)\in[0,1]$ is the normalized effort worker $i$ spends during time slot $t$. $m_{i}(t)\in[0,1]$ is worker $i$'s mood during time slot $t$, where 1 denotes the most positive mood. It can be self-reported by a worker using tools such as \textit{MoodPanda} (\url{http://moodpanda.com/}), or through facial image analytics (e.g., in-vehicle monitoring of drivers' mood \cite{Zimasa-et-al:2017}). Here, we treat $m_{i}(t)$ as an external parameter to CPL and do not assume that the system is capable of predicting future mood values.

\subsection{Deriving the Work-Rest Index}
In order to take task pending time into account, we propose a \textit{conceptual queue} technique. Let $Q_{i}(t)$ denote a conceptual queue which is being managed by a CPL agent on behalf of a worker $i$. The queuing dynamics of this conceptual queue is defined as:
\begin{equation}
Q_{i}(t+1)=\max[0,Q_{i}(t)+\mu_{i}^{\max}1_{[q_{i}(t)>0\text{ \& }\mu_{i}(t)=0]}-\mu_{i}(t)] \label{eq:vQ}
\end{equation}
where $1_{[\textnormal{condition}]}$ is an indicator function. Its value is 1 if and only if [condition] is satisfied; otherwise, it evaluates to 0. When first created, a conceptual queue is empty. The conceptual queue starts from 0 at $t=0$ (i.e. $Q_{i}(0)=0$). Its size increases by $\mu_{i}^{\max}$ if and only if worker $i$'s pending task queue is not empty at time slot $t$ and worker $i$ rests during this time slot. Its size decreases in the same way as $q_{i}(t)$.

For simplicity of notation, we denote $\mu_{i}^{\max}1_{[q_{i}(t)>0\text{ \& }\mu_{i}(t)=0]}$ as $x_{i}(t)$. Then, equation (\ref{eq:vQ}) can be re-expressed as:
\begin{equation}
Q_{i}(t+1)\geqslant Q_{i}(t)+x_{i}(t)-\mu_{i}(t).
\end{equation}
By re-arranging the above inequality, we have:
\begin{equation}
Q_{i}(t+1)-Q_{i}(t)\geqslant x_{i}(t)-\mu_{i}(t).
\end{equation}
Summing both sides of the above inequality over all $t\in\{0,...,T-1\}$ yields:
\begin{equation}
\sum_{t=0}^{T-1}[Q_{i}(t+1)-Q_{i}(t)]\geqslant \sum_{t=0}^{T-1}[x_{i}(t)-\mu_{i}(t)].
\end{equation}
Thus, we have:
\begin{equation}
Q_{i}(T)-Q_{i}(0)\geqslant \sum_{t=0}^{T-1}[x_{i}(t)-\mu_{i}(t)].
\end{equation}
Since $Q_{i}(0)=0$, the above inequality is simplified as:
\begin{equation}
\frac{Q_{i}(T)}{T}\geqslant \frac{1}{T}\sum_{t=0}^{T-1}x_{i}(t)-\frac{1}{T}\sum_{t=0}^{T-1}\mu_{i}(t).\label{eq:vQ2}
\end{equation}

From equation (\ref{eq:vQ2}), it can be observed that the effect of the conceptual queue is to signal the scheduling approach as to when the need to reduce pending workload shall take precedence over helping workers plan their rest. By ensuring that the computed $\mu_{i}(t)$ values satisfy the queueing stability requirement of $\frac{1}{T}\sum_{t=0}^{T-1}\mu_{i}(t)\geqslant\frac{1}{T}\sum_{t=0}^{T-1}x_{i}(t)$ for the conceptual queue, a scheduling approach will ensure that tasks do not stay pending indefinitely.

By jointly considering $q_{i}(t)$ and $Q_{i}(t)$, the \textit{Lyapunov function} \cite{Neely:2010} which measures the overall concentration of work among workers in a given system during time slot $t$ is:
\begin{equation}
L(t)=\frac{1}{2}\sum_{i=1}^{N}[q_{i}^{2}(t)+Q_{i}^{2}(t)]. \label{eq:L}
\end{equation}
A large value of $L(t)$ indicates that tasks are highly concentrated among a small number of workers and/or that tasks have been pending for a long period of time. Both of these scenarios are undesirable from a system productivity perspective and shall be avoided as much as possible. The constant term $\frac{1}{2}$ is included to simplify subsequent derivations without affecting the physical meaning of the formulation. We adopt the time-averaged \textit{Lyapunov drift}, $\Delta=\frac{1}{T}\sum_{t=0}^{T-1}[L(t+1)-L(t)]$, to measure the changes in the degree of seriousness of these two scenarios over time.

We formulate a joint \textit{\{effort output + drift\}} optimization objective function as:
\begin{equation}
\phi\mathbb{E}\{\tilde{\xi}(t)|\tilde{q}(t),\tilde{m}(t)\}+\Delta \label{eq:obj}
\end{equation}
which shall be minimized. $\phi>0$ controls the emphasis placed on conserving worker effort compared to getting more work done. A larger $\phi$ value can be interpreted as stronger emphasis on allowing workers to rest more. This value can be set by the system operators to express system-level preference on how to utilize the collective productivity of the workers. $\tilde{\xi}(t)$, $\tilde{q}(t)$ and $\tilde{m}(t)$ are vectors containing the workers' effort output values, the backlog queue sizes, and the self-reported mood during time slot $t$, respectively.

Based on equation (\ref{eq:Q}) and equation (\ref{eq:L}), the time-averaged Lyapunov drift can be expressed as:
\begin{dmath}
\Delta=\frac{1}{T}\sum_{t=0}^{T-1}\sum_{i=1}^{N}\left[\left(\frac{1}{2}q_{i}^{2}(t+1)-\frac{1}{2}q_{i}^{2}(t)\right)+\left(\frac{1}{2}Q_{i}^{2}(t+1)-\frac{1}{2}Q_{i}^{2}(t)\right)\right]\\
=\frac{1}{T}\sum_{t=0}^{T-1}\sum_{i=1}^{N}\left[\left(\frac{1}{2}\max[0,q_{i}(t)+\lambda_{i}(t)-\mu_{i}(t)]^{2}-\frac{1}{2}q_{i}^{2}(t)\right)+\left(\frac{1}{2}\max[0,Q_{i}(t)+\mu_{i}^{\max}1_{[q_{i}(t)>0\text{ \& }\mu_{i}(t)=0]}-\mu_{i}(t)]^{2}-\frac{1}{2}Q_{i}^{2}(t)\right)\right]\\
\leqslant\frac{1}{T}\sum_{t=0}^{T-1}\sum_{i=1}^{N}\left[\left(q_{i}(t)[\lambda_{i}(t)-\mu_{i}(t)]-\mu_{i}(t)\lambda_{i}(t)+\frac{1}{2}[\lambda_{i}^{2}(t)-2\lambda_{i}(t)\mu_{i}(t)+\mu_{i}^{2}(t)]\right)\\
+\left(Q_{i}(t)[\mu_{i}^{\max}1_{[q_{i}(t)>0\text{ \& }\mu_{i}(t)=0]}-\mu_{i}(t)]+\frac{1}{2}[(\mu_{i}^{\max})^{2}1_{[q_{i}(t)>0\text{ \& }\mu_{i}(t)=0]}-2\mu_{i}^{\max}1_{[q_{i}(t)>0\text{ \& }\mu_{i}(t)=0]}\mu_{i}(t)+\mu_{i}^{2}(t)]\right)\right].
\end{dmath}
This formulation enables simultaneous modelling of the absolute sizes of the real and the conceptual queues, the distribution of congestions among these queues, and the fluctuations of these queues over time. All three quantities should be minimized in order to optimize our design objectives. In this way, we translate system-level efficiency and worker wellbeing requirements into queueing system stability concepts. Then, through maintaining queue system stability, CPL achieves these desirable objectives.

Since neither $\lambda_{i}(t)$ nor $\mu_{i}(t)$ can be infinite in real-world systems, we can simplify $\Delta$ as:
\begin{dmath}
\Delta\leqslant\frac{1}{T}\sum_{t=0}^{T-1}\sum_{i=1}^{N}\left[\left(q_{i}(t)[\lambda_{i}(t)-\mu_{i}(t)]-\mu_{i}(t)\lambda_{i}(t)+\frac{1}{2}[\lambda_{\max}^{2}+\mu_{\max}^{2}]\right)\\
+\left(Q_{i}(t)[\mu_{\max}1_{[q_{i}(t)>0\text{ \& }\mu_{i}(t)=0]}-\mu_{i}(t)]+\frac{1}{2}[\mu_{\max}^{2}1_{[q_{i}(t)>0\text{ \& }\mu_{i}(t)=0]}+\mu_{\max}^{2}]\right)\right]. \label{eq:surprise}
\end{dmath}
where $\lambda_{\max}\geqslant\lambda_{i}(t)$ and $\mu_{\max}\geqslant\mu_{i}(t)$ for all $i$ and $t$ are constant values. As we only aim to influence $\mu_{i}(t)$ with recommendations, we only retain terms containing $\mu_{i}(t)$. In this way, by substituting equation (\ref{eq:surprise}) into equation (\ref{eq:obj}), we obtain the following objective function:

 Minimize:
\begin{equation}
\frac{1}{T}\sum_{t=0}^{T-1}\sum_{i=1}^{N}\xi_{i}(t)[\phi-(q_{i}(t)+Q_{i}(t))m_{i}(t)\mu_{i}^{\max}] \label{eq:obj1}
\end{equation}

 Subject to:
\begin{equation}
0\leqslant \xi_{i}(t)\leqslant 1, \forall i, \forall t \label{const1-1}
\end{equation}
\begin{equation}
0\leqslant\mu(\xi_{i}(t),m_{i}(t))\leqslant\mu_{i}^{\max}, \forall i, \forall t \label{const1-2}
\end{equation}
By minimizing equation (\ref{eq:obj1}) subject to Constraints (\ref{const1-1}) and (\ref{const1-2}), we simultaneously minimize the time-averaged total worker effort output while maximizing the time-averaged collective productivity in a given system. To minimize equation (\ref{eq:obj1}), at each time slot $t$, we need to compute the values of the expression $[\phi-(q_{i}(t)+Q_{i}(t))m_{i}(t)\mu_{i}^{\max}]$ for all $i$. For simplicity of notation, we denote $[\phi-(q_{i}(t)+Q_{i}(t))m_{i}(t)\mu_{i}^{\max}]$ as the \textit{Work-Rest Index (WRI)}, $\Psi_{i}(t)$. This index enables a scheduling agent to efficiently search through a very large solution space to determine if the current situation is more suitable for work or rest.

\subsection{Opportunistic Work-Rest Scheduling}
Algorithm \ref{al:1} presents a distributed implementation of the CPL approach as a scheduling agent for each worker. For a worker $i$, if $\Psi_{i}(t)<0$, it sets $\xi_{i}(t)=\min\left[1,\frac{q_{i}(t)}{m_{i}(t)\mu_{i}^{\max}}\right]$ and computes the corresponding $\mu_{i}(t)$ value; otherwise, it advises the worker to rest for the current time slot. The computational time complexity of Algorithm \ref{al:1} is $O(1)$, making it highly scalable. The algorithm implements computationally the intuition that \textit{the more pending tasks in a worker's backlog queue, the longer these tasks have been pending, the more emphasis on worker wellbeing by system operators, and the higher the worker's current mood, the more effort should be expended towards completing tasks (subject to the physical limitations of the worker's effort output per time slot)}.

\begin{algorithm}[ht]
\caption{CPL}
\begin{algorithmic}[1] \label{al:1}
\REQUIRE $\phi$, $q_{i}(t)$, $\mu_{i}^{\max}$ and $m_{i}(t)$.
\IF {$\Psi_{i}(t)<0$}
    \STATE $\xi_{i}(t)=\min\left[1,\frac{q_{i}(t)}{m_{i}(t)\mu_{i}^{\max}}\right]$;
    \STATE Compute $\mu_{i}(t)$ according to equation (\ref{eq:mu1});
\ELSE
    \STATE $\xi_{i}(t)=0$;
    \STATE $\mu_{i}(t)=0$;
\ENDIF
\RETURN $\mu_{i}(t)$;
\end{algorithmic}
\end{algorithm}

WRI incorporates the ethical considerations \cite{Yu-et-al:2018IJCAI} by prioritizing worker wellbeing considerations through recommending opportunistic rest breaks, and allowing stakeholders to influence the AI recommendations by expressing their preferences through $\phi$ and $m_{i}(t)$. The recommendations from CPL are given to a worker in the form of the number of tasks he should complete over a given time slot (a 0 value indicates that the worker should rest) so as to make it actionable enough for the worker to follow.

\section{Experimental Evaluation}
To evaluate the performance of CPL under realistic settings, we compare it against four alternative approaches through extensive simulations. The characteristics of worker agents in the simulation are derived from the \textit{Tianchi} dataset (\url{http://dx.doi.org/10.7303/syn7373599}) released by Alibaba. This real-world dataset contains information regarding 5,547 workers' reputation (i.e. quality of work) and maximum productivity (i.e. the maximum number of tasks a worker can complete per time slot). This dataset allows us to construct realistic simulations.

\subsection{Experiment Settings}
The five comparison approaches are:
\begin{enumerate}
    \item The \textit{Max Effort (ME)} approach: under this approach, a worker agent $i$ always works as long as there are tasks in its backlog queue regardless of its mood.
    \item The \textit{Mood Threshold (MT)} approach: under this approach, a worker agent $i$ works whenever $m_{i}(t)\geqslant \theta_{1}$ and there are tasks in its backlog queue, where $\theta_{1}\in[0,1]$ is a predetermined mood threshold used by MT.
    \item The \textit{Mood and Workload threshold (MW)} approach: this approach jointly considers a worker agent $i$'s current mood and workload to determine how much effort to exert. Whenever $q_{i}(t)\mu(1,m_{i}(t))\geqslant \mu_{i}^{\max}\mu(1,\theta_{2})$, worker $i$ exerts up to the maximum effort subject to there being enough tasks in its backlog queue, where $\theta_{2}\in[0,1]$ is a predetermined mood threshold used by MW.
    \item The \textit{Affective Crowdsourcing (AC)} approach \cite{Yu-et-al:2017ICA} which is similar to CPL but does not take task pending time into account.
    \item The \textit{CPL} approach proposed in this paper.
\end{enumerate}

The approach used to delegate tasks to worker agents under all comparison approaches is SMVM \cite{Yu-et-al:2017SciRep}. It dynamically distributes tasks among workers in a situation-aware manner in order to avoid over concentration of workload. At each time slot, SMVM determines how many tasks to delegate to each worker agent $i$ in the simulations (i.e. SMVM computes $\lambda_{i}(t)$ for all $i$ and $t$) based on its current reputation and workload. The principle implemented by SMVM is that the higher a worker agent's reputation and the lower its current workload, the more tasks should be delegated to it. SMVM can be replaced by any other approach \cite{Ho-et-al:2013,BasuRoy-et-al:2015} as long as such an approach can determine the values of $\lambda_{i}(t)$ for all $i$ and $t$.

In order to create different experiment conditions, we vary the value of $\phi$ between 5 and 100 in increments of 5. The values of $\theta_{1}$ and $\theta_{2}$  are varied between 0.05 and 1 in increments of 0.05. The system workload is measured in relation to the maximum collective productivity of the worker agent population, $\Omega=\sum_{i=1}^{N}r_{i}\mu_{i}^{\max}$. In this equation, $r_{i}$ is a worker agent $i$'s reputation and $N=5,547$. We adopt the concept of load factor (LF) from \cite{Yu-et-al:2016SciRep,Yu-et-al:2017SciRep} to denote the overall workload placed on the system. It is computed as the ratio between the number of new tasks delegated to the worker agents during time slot $t$, $W_{req}(t)$, and the maximum collective productivity $\Omega$ of the system (i.e. $LF=\frac{W_{req}(t)}{\Omega}$). We vary LF between 5\% to 100\% in 5\% increments.

Throughout the experiments, the mood for each worker agent $i$ during time slot $t$, $m_{i}(t)$, is randomly generated in the range of $[0,1]$ following a uniform distribution. This eliminates the possibility for any of the comparison approaches to predict a worker agent's future mood based on previous observations, thereby focusing the experimental comparisons on the effectiveness of the scheduling strategies. Under each LF setting, the simulation is run for $T=10,000$ time slots. All tasks must be completed within 3 time slots after they have been delegated.

\subsection{Evaluation Metrics}
The performances of the five approaches in the experiments are compared using the following metrics:
\begin{enumerate}
    \item The time-averaged worker effort output, $\bar{\xi}=\frac{1}{TN}\sum_{t=0}^{T-1}\sum_{i=1}^{N}\xi_{i}(t)$. The smaller the $\bar{\xi}$ value, the better the performance of an approach.
    \item The time-averaged task expiry rate, $\bar{e}=\frac{1}{TN}\sum_{t=0}^{T-1}\sum_{i=1}^{N}\frac{n_{i}^{(e)}(t)}{q_{i}(t)}$, where $n_{i}^{(e)}(t)$ is the number of tasks in $i$'s backlog which passed their deadlines during a given time slot $t$. The smaller the $\bar{e}$ value, the better the performance of an approach.
    \item The time-averaged task completion rate, $\bar{\mu}=\frac{1}{T}\sum_{t=0}^{T-1}\frac{\sum_{i=1}^{N}\mu_{i}(t)}{N_{\text{total}}(t)}$, where $N_{\text{total}}(t)$ is the total number of tasks waiting to be completed in the system during a given time slot $t$. The larger the $\bar{\mu}$ value, the better the performance of an approach.
\end{enumerate}

Since worker agents under the ME approach consistently expend the most effort and achieve the highest task completion rate, we use ME as the baseline for comparing the performance of other approaches under different LF, $\phi$, $\sigma$, $\theta_{1}$ and $\theta_{2}$ settings. 

\subsection{Results and Discussions}
\begin{figure*}[ht]
    \centering
    \includegraphics[trim = 40mm 0mm 30mm 10mm, clip, width=1\linewidth]{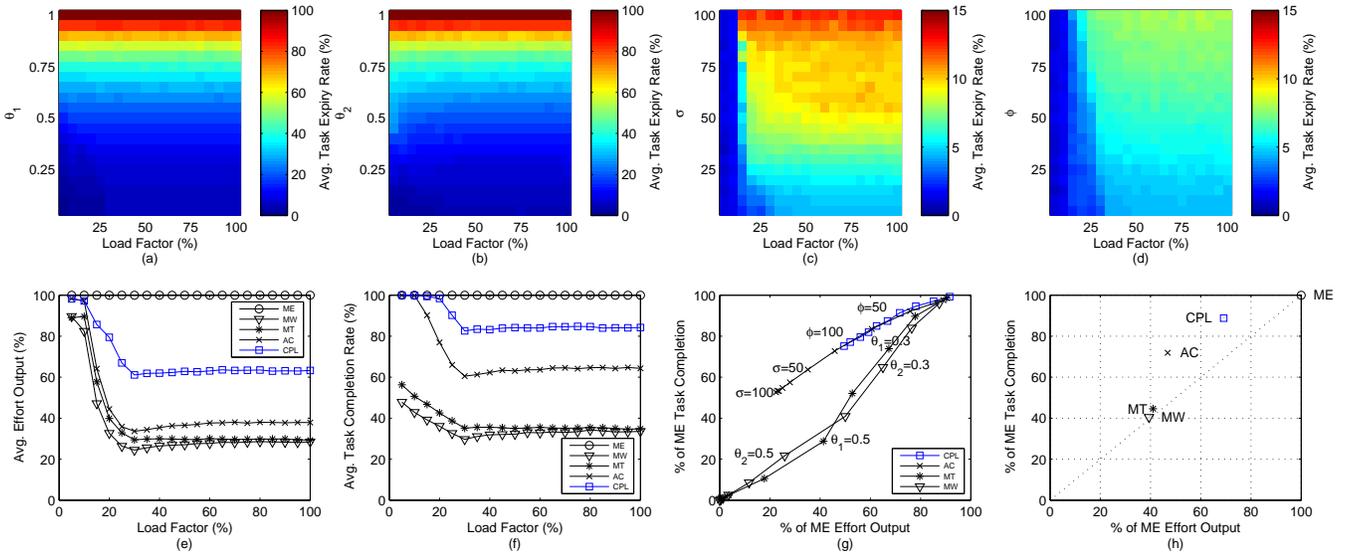}
    \vspace*{-12pt}
    \caption{Experiment results: {\bf(a)} the time-averaged task expiry rates achieved by MT under various $\theta_{1}$ and LF settings; {\bf(b)} the time-averaged task expiry rates achieved by MW under various $\theta_{2}$ and LF settings; {\bf(c)} the time-averaged task expiry rates achieved by AC under various $\sigma$ and LF settings; {\bf(d)} the time-averaged task expiry rates achieved by CPL under various $\phi$ and LF settings; {\bf(e)} comparison of the time-averaged effort output achieved by various approaches under different LF settings; {\bf(f)} comparison of the time-averaged task completion rates achieved by various approaches under different LF settings; {\bf(g)} the time-averaged task completion rates vs. the time-averaged effort output achieved by various approaches under different parameter settings as a percentage of those achieved by ME; {\bf(h)} the time-averaged task completion rate vs. the time-averaged effort output achieved by various approaches averaged over different parameter settings as a percentage of those achieved by ME.}\label{fig:Results}
\end{figure*}

Figures \ref{fig:Results}(a)--\ref{fig:Results}(d) show the time-averaged task expiry rates achieved by MT, MW, AC and CPL respectively under various experiment settings. As MT uses mood as the threshold to control worker effort output, the changes in task expiry are directly related to mood values (Figure \ref{fig:Results}(a)). On average, 29\% of the tasks under MT expire before they can be completed. MW is also a threshold-based approach. However, its threshold consists of a combination of workers' mood and their current workload. Thus, its task expiry rate increases with both mood and LF with the effect of mood being more significant (Figure \ref{fig:Results}(b)). On average, 30\% of the tasks under MW expire before they can be completed. AC is not a threshold-based scheduling approach. A worker can indicate to AC his desire to rest by adjusting the value of the control variable $\sigma$. As $\sigma$ and LF values increase, an increasing percentage of tasks expire under AC. On average, 7.5\% of the tasks under AC expire before they can be completed, which is significantly lower than MT and MW. As AC only considers mood and workload when making work-rest recommendations, workers may fall into the condition in which their mood and workload trigger AC to recommend resting. However, their workload is also not low enough to cause the task delegation approach to delegate new tasks to them. Therefore, AC continues to recommend resting until pending tasks pass their deadlines and become expired. This problem is addressed by CPL as it takes task pending time into account with the conceptual queue technique when optimizing work-rest scheduling. As shown in Figure \ref{fig:Results}(d), increases in $\phi$ and LF values result in an increasing percentage of tasks expire under CPL. The task expiry rate under CPL is lower than that under AC. On average, 5.4\% of the tasks under CPL expire before they can be completed, which is significantly lower than MT, MW and AC considering the scale of the experiment.

The time-averaged effort output $\bar{\xi}$ achieved by all five approaches is shown in Figure \ref{fig:Results}(e). Compared to ME, all other approaches achieved significant savings in effort as LF increases. This is partially due to the SMVM task delegation approach used in the simulations. When LF is low, tasks are mostly concentrated on worker agents with good reputation. In this case, the task backlogs of individual worker agents who have been delegated tasks tend to be relatively high, which makes scheduling approaches allocate less time for these workers to rest in order to meet task deadlines. As LF increases, the workload is spread more evenly among a larger segment of the worker agent population, creating more opportunities for scheduling approaches to slot in rest breaks. The $\bar{\xi}$ values achieved by MT, MW and AC stabilize between 20\% and 40\% while that of CPL stabilizes around 60\%. MW achieves the lowest worker agent effort output. The time-averaged task completion rates $\bar{\mu}$ achieved by all five approaches are shown in Figure \ref{fig:Results}(f). It can be observed that CPL achieves the highest $\bar{\mu}$ values which stabilize around 85\% and are higher than AC, MT and MW by 22\%, 90\% and 108\%, respectively.

Figure \ref{fig:Results}(g) shows the performance landscape of MT, MW, AC and CPL as a percentage of ME under different control parameter settings. MW and MT both use mood thresholds ($\theta_{1}$ and $\theta_{2}$, respectively) to control effort output. The higher the mood threshold values, the lower the effort output (and hence the task completion rates) achieved by MW and MT. On the other hand, mood serves as one of the inputs to AC and CPL. Both AC and CPL allow a worker agent to specify a general emphasis on conserving effort using variables $\sigma$ and $\phi$, respectively. The larger the values of these variables, the more emphasis is placed on effort conservation. By varying the $\phi$ value in CPL, we can control the trade-off between worker effort output and task completion rate (from spending 92\% of the ME effort output and achieving 99\% of the ME task completion rate, to spending 53\% of the ME effort output and achieving 78\% of the ME task completion rate). CPL consistently and significantly outperforms both MT and MW, conserving significant worker effort while achieving high task throughput. In effect, CPL limits the range of trade-off between work and rest achieved by AC based on consideration of an additional situational factor -- the task pending time -- in order to achieve better collective performance. In the worst case scenario in which $\theta_{1}$ and $\theta_{2}$ are set to 1, indicating that workers are unwilling to work under any mood condition, the worker effort output and the task completion rates achieved by both MT and MW are 0. This is expected as in MT and MW, mood is used as the threshold to control effort output. However, under AC, mood is only one of the situational factors considered by the approach. CPL adds in task pending time on top of the situational factors used by AC to make scheduling decisions. Even under scenarios in which $\phi$ is set to 100 (indicating that workers place very high emphasis on rest), CPL is still able to maintain a long-term average effort output of 53\% taking advantage of favourable working conditions whenever possible to achieve average task completion rates of about 78\%. This is more advantageous from a business perspective compared to the worst case performance by AC of about 50\%.

When we compute the averages of the results shown in Figure \ref{fig:Results}(g) over their respective setting variables (i.e. $\phi$, $\sigma$, $\theta_{1}$ and $\theta_{2}$), we obtain Figure \ref{fig:Results}(h) showing the overview of their performances. The diagonal dotted line represents linear productivity, meaning that an increase in effort output results in a directly proportional increase in collective productivity. It can be observed that ME, MT and MW all fall on the linear productivity line, whereas AC and CPL are significantly above this line in a region of superlinear productivity \cite{Sornette-et-al:2014}. Under AC and CPL, an increase in effort output results in a disproportionally larger increase in collective productivity, indicating that the collective productivity achieved is larger than the sum of individual workers' productivity. Overall, CPL achieves 89\% of ME task completion rate with 69\% of the ME worker effort output, which is the most desirable work-rest trade-off among the five approaches from a system perspective.

\section{Conclusions and Future Work}
Improving collective productivity is an important problem facing many social and economic systems. How to dynamically adapt workers' work-rest schedules in response to changing situations in order to maintain a high level of productivity and worker wellbeing remains an open research question. The proposed CPL approach translates considerations on workers' mood, workload and pending time of the tasks in their backlogs into actionable personalized work-rest schedules. It establishes a framework to model complex relationships between work and rest, and helps workers optimize the balance between work and rest in order to achieve superlinear collective productivity. Taking into account its polynomial time complexity, CPL is an effective and scalable approach to help workers benefit from opportunistic rest. By nudging workers to be `lazy' at opportune times, CPL achieves collective productivity which is larger than the sum of individual productivity. It provides a way to design ethically aligned workforce management systems that sustain long-term effective participation by promoting productive laziness among workers.

In future research, we plan to testbed CPL in a crowdsourcing platform \cite{Pan-et-al:2016AAAI} to reach out to more diverse users and study how to improve the approach in the presence of various behaviour patterns and how to foster trust \cite{Shen-et-al:2011} with user by explaining the rationale behind the recommendations.

\small
\section*{Acknowledgments}
This research is supported, in part, by Nanyang Technological University, Nanyang Assistant Professorship (NAP) and the National Research Foundation, Prime Minister's Office, Singapore under its IDM Futures Funding Initiative.

\bibliographystyle{aaai}

\end{document}